# APPLICATION OF ARTIFICIAL INTELLIGENCE IN THE CLASSIFICATION OF MICROSCOPICAL STARCH IMAGES FOR DRUG FORMULATION

Marvellous Ajala, Blessing Oko, David Oba-Fidelis, Joycelyn Iyasele, Joy I. Odimegwu

Department of Pharmacognosy, Faculty of Pharmacy, University of Lagos, Nigeria.

**ABSTRACT**

Starches are important energy sources found in plants with many uses in the pharmaceutical industry such as binders, disintegrants, bulking agents in drugs and thus require very careful physicochemical analysis for proper identification and verification which includes microscopy.

In this work, we applied artificial intelligence techniques (using transfer learning and deep convolution neural network CNNs to microscopical images obtained from 9 starch samples of different botanical sources. Our approach obtained an accuracy of 61% when the machine learning model was pretrained on microscopic images from MicroNet dataset. However the accuracy jumped to 81% for model pretrained on random day to day images obtained from Imagenet dataset. The model pretrained on the imagenet dataset also showed a better precision, recall and f1 score than that pretrained on the imagenet dataset.

# INTRODUCTION

Starch is one of the most widely and easily available polymers present in nature and can be obtained from various fruits and vegetables. It is also commonly used as food in our daily life (Bordoloi et al., 2019). Starch is also the most abundant storage of carbohydrates in plants and a naturally occurring polysaccharide whose distribution is only second to cellulose as the most abundant biomass material found in nature (Emeje et al, 2012). Carbohydrates are produced as a result of photosynthetic activity responsible for storing the chemical energy of the sun in different parts of plants including the leaves of green plants, seeds, fruits, stems, roots, and tubers of most plants and making it available to non-photosynthetic organisms with humans being the most significant beneficiaries.

Starch is structurally a polysaccharide that is composed of monosaccharide units (glucose) connected by α-D-(1-4) and α-D-(1-6) linkages. The two major types of polymers found in starch are amylose and amylopectin. Amylose refers to the linear chain of α-1,4 glycosidic bonds with limited branching points at the α-1,6 positions and makes up about 15 and 30% of common starch. Amylopectin however consists of linear chains of glucose units linked by α-1,4 glycosidic bonds but also highly branched at the α-1,6 positions by small glucose chains at intervals of 10 nm along the molecule's axis, constituting the bulk of starch known today (about 70–85% of common starch).

But beyond their natural importance as food, starches have several uses in the pharmaceutical industry due to their different physical properties such as the structure of the starch granule (Singh et al, 2003), swelling capacity (Conde-Petit et al, 2001), solubility (Singh et al 2003), gelatinization (Jane, 1995), retrodation (Singh et al, 2003) etc. Thus, based on the different properties found in different starches, they are used for different purposes as pharmaceutical excipients for drug production. For example, due to the hydrophilic nature of some starches, maize and potato starches are commonly used as disintegrating agents in chloroquine tablets (Pilpel et al, 1978), and starch paste from corn starch is commonly used as a binder due to its gelling properties in paracetamol tablet (Akin-Ajadi et al, 2005), starch is used as film forming coating due to the film forming property of one of its major constituents, amylose (Myllarinen et al, 2002) etc.

Thus, it is also important to avoid mixups of the starch samples as this can be very fatal and expensive for the pharmaceutical company and formulating scientist, for example using of disintegrating agent instead of a binder might disrupt the physicochemical property of the drug formulation

Due to the different uses of starch obtained from different botanical origins, it is therefore necessary and important that identification and quality analysis is carried out on the starch sample during quality assurance and control, before and after incorporation into the drug formulation. These physicochemical tests include organoleptic tests, bulk and tapped density, compressibility index and Hausner ratio, angle of repose (USP 40-NF 35), solubility and swelling powers (Walter et al, 2000), amylose-amylopectin ratio (Wang et al, 2010), microscopy such as Scanning Electron Microscopy (SEM), Fourier Transform and Infrared Spectroscopy (FT-IR) (Azeh et al 2018) and chemical tests such as Toludiene test ( Bisulca et al, 2016) among many others.

These procedures are however often labour-intensive, time-consuming and require technical knowledge available to just a few specialised individuals. Also, misidentification of starch could occur due to human error.

Machine Learning is defined by Samuel Arthur in 1959 as a field of study that gives computers the ability to learn without being explicitly programmed (Park et al., 2018). It is also defined as "A computer program is said to learn from experience E with respect to some task T and some performance measure P, if its performance on T, as measured by P, improves with experience E". (Mitchell TM, 1997).

Machine learning is thus a field of study where computers are trained to solve tasks which they were not explicitly programmed to solve but rather by showing examples. This field was developed to solve challenges that were originally difficult to explicitly program and also to cater for the ever-increasing quantity of data being produced continually. As a machine learning algorithm gets more experience, by being exposed to observational data or interactions with an environment, its performance improves.

Machine Learning can then be further divided into supervised and unsupervised learning (Géron, 2019).

In supervised learning, the machine learning system is provided with data that contain both the examples (features) and their appropriate label and the system is then taught to learn the relationship by mathematically mapping the features to the label. Tasks that can then be done under supervised learning include classification tasks, where the model is taught to categorize things into the respective classes, regression tasks where the model predicts a continuous value given certain features, e.g cost of an instrument given the description of the instruments, tagging where the features are appended with all correct tags/labels, search where a system is taught to return specific values when given a query, recommendation where the system gives suggestions on probable items given some previous history etc.

Unsupervised learning on the other hand involves the machine learning system learning patterns within the dataset given to it without any label given to guide the exploration. Because of the nature of this system, tasks associated with unsupervised learning are therefore aimed towards generally understanding the dataset more without the context of any label, some of those tasks involve clustering analysis, to identify existing clusters or groups of similar items in the dataset, dimensionality reduction to reduce the dimensions (number of features) of a given data to a smaller but informative dimension and visualization tasks to visualize and understand the datasets.

There also exists reinforcement learning where the system is taught to engage/play against itself with a reward and a penalty system for desirable and undesirable behaviors. An example of this is Alphago (Silver et al., 2016), the machine learning system that learnt to play chess by playing against itself.

Broadly defined, Artificial Intelligence encompasses many branches of statistical and machine learning, pattern recognition, clustering, similarity-based methods, logic and probability theory, as well as biologically motivated approaches, such as neural networks and fuzzy modeling that information scribed as "computational intelligence". The emergence and success of machine learning during the Imagenet competition (Russakovsky et al., 2015) have led to the adoption of machine learning in different aspects and fields such as computer vision, natural language processing, automatic speech recognition, natural language understanding and also in niche fields like healthcare and pharmaceuticals.

**OBJECTIVE:**

In this research, we seek to apply machine learning techniques to identify starch of different origins by using their microscopical images

**METHODOLOGY**

**Dataset**

The dataset consists of 889 pictures from 9 different starches namely the starch of the kernel of Maize (*Zea mays, Linn), the* starch of Fresh Tubers of Cassava (*Manihot utilissima* Pohl.), the starch of root tuber of Irish Potato (*Solanum tuberosum* Linn.), starch of the grain of wheat (*Triticum aestivum*), starch of the grain of rice ( *Oryza sativa), the* starch of the grain of oat (*Avena sativa*), the starch of the peas of green pea (*Pisum sativum* L.), *the* starch of the tuber of Tiger nut (*Cyperus esculentus* L.) and the starch of the grain of millet (*Pennisetum glaucum* L.)

**Table 1. Table of the different starches and number of images used**

| S/N | STARCH | NUMBER |
|---|---|---|
| 1 | Cassava starch | 110 |
| 2 | Green peas starch | 119 |
| 3 | Irish Potato starch | 100 |
| 4 | Maize starch | 100 |
| 5 | Millet starch | 103 |
| 6 | Oat starch | 81 |
| 7 | Rice starch | 91 |
| 8 | Wheat starch | 112 |
| 9 | Tiger nut starch | 83 |
|  | Total | 889 |

Microscopic slides of the starches were prepared and they were mounted using glycerol. The starch granules were studied under a light microscope. Images of the microscopical view of the starch samples were taken with the aid of a phone camera at x640 magnification.

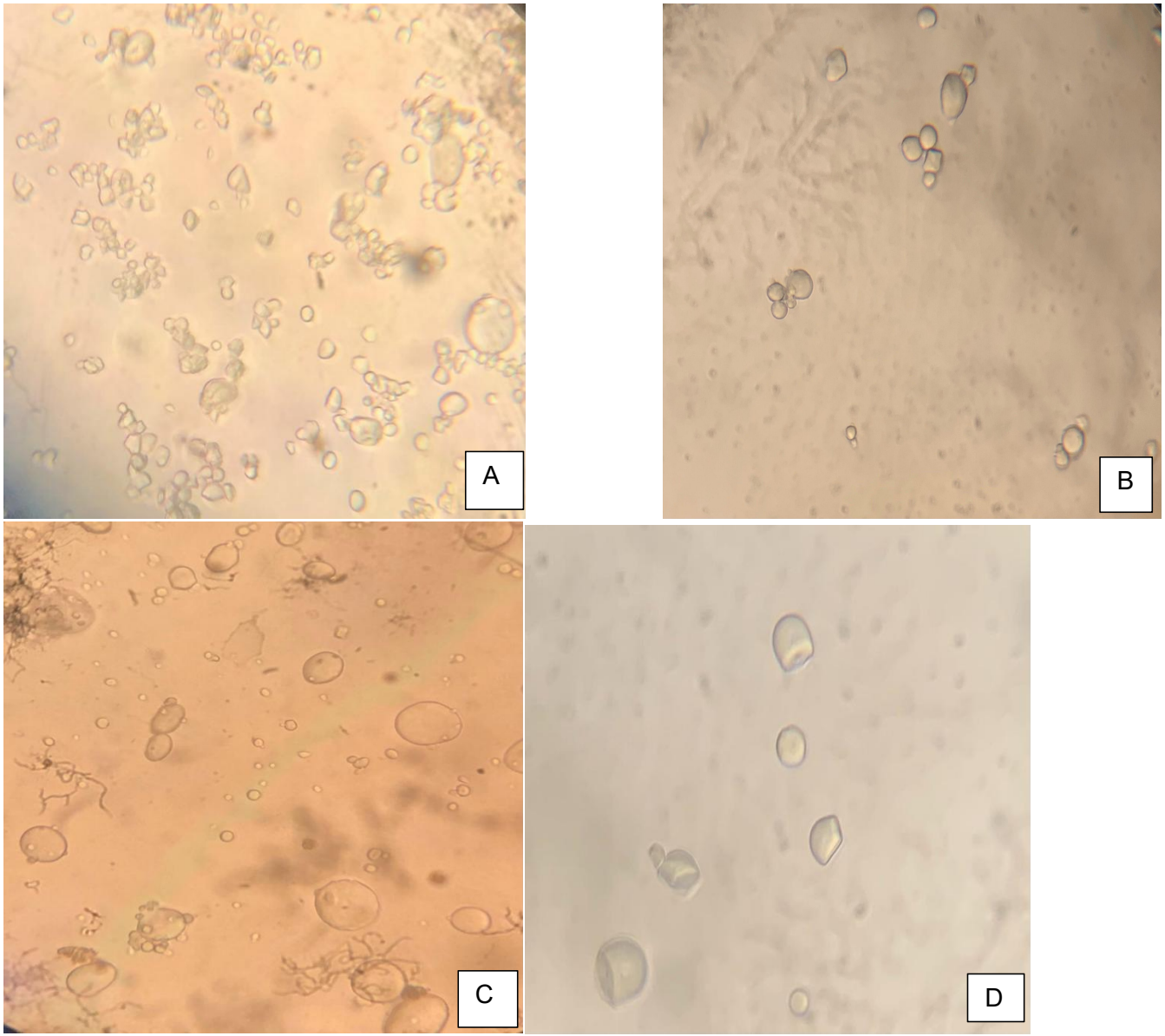

Fig 1. Examples of microscopical images of the starches: from left to right, the image of cassava starch [A], the image of maize starch [B], the image of oat starch [C] and the image of wheat starch [D] Taken with light microscope at Mag X640

**Data Pre- processing and augmentation.**

All the pictures were taken in ".jpeg" or ".heic" picture formats. The pictures in ".heic" were all then converted to ".jpeg" format using heic-to-jpg python package (Reimers, 2020) before any further processing was done on any of the images. Using torchvision (Marcel & Rodriguez, 2010), the images were all resized to 224 by 224 pixels to ensure the uniformity of all the pictures. The image values were also normalized with a mean of 0.5 and a standard deviation of 0.5 to reduce the skewness of the data and increase the speed of processing. Some of the images were randomly flipped horizontally and also randomly rotated by about $10^0$.

**Model design**

For the models, 2 pretrained ResNet models ( Szegedy et al., 2017) were trained and evaluated, one pretrained on Micronet dataset (Stuckner et al., 2022) that contains over 100, 000 microscopical images and the other was pretrained on Imagenet dataset (Deng et al, 2009), which contain 14 million pictures of Ron random day to day activities . The models output layers were replaced by 3 fully connected layers having  500, 100 and 9 hidden nodes, ReLU activation function (Agarap, 2018) and 0.5 Dropout layer (Srivastava et al., 2014). The models were built with PyTorch (Paszke et al., 2019).

The ResNet model belongs to a class of deep learning (a subset of machine learning that was inspired and modelled similarly to the neurons in the brain (Poo, 2018)) that are adapted specifically for computer vision tasks known as Convolution Neural Networks (CNNs/ConvNets) and became the de facto choice for image classification tasks (Raghu et al., 2021) following the Imagenet competition (Krizhevsky et al., 2017). They receive as input images and consist of convolving filters (convolutions) that can extract features of pictures by applying convolutions defined by mathematical functions to multiple sections of an image to capture the local invariance and locality in images. They then apply activation functions to capture nonlinear relationships in the image, and pooling functions to reduce the picture dimension space while retaining the most important information.

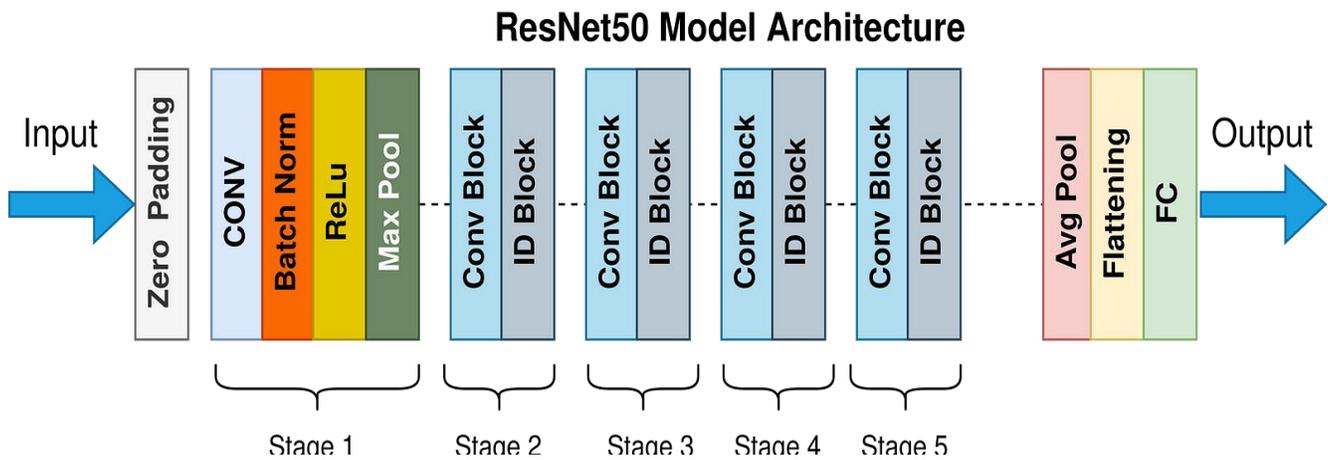

Fig 2. Resnet18 model architecture

**2.4 Training**

The data was split into 3 sets, a training set (50% of the data), a test set (20% of the data) and a validation set (20% of the remaining data).

The model used the Adams optimizer and also had an early stoppage function which stopped the model from training any further after not improving for 5 iterations.

The model had an epoch size of 30, batch size of 8 images, a learning rate of 0.001, NNLogloss as loss function and validation loss as evaluation criteria. During the training, for the original ResNet model, the model weights were downloaded from the PyTorch hub, while the MicoNet model weights were also downloaded from the provided URL (uniform resource locator). The models were trained on the trainset and evaluated on the test_set. After training, the overall best checkpoint was loaded and evaluated on the validation dataset for both the original ResNet and Micronet models. The accuracy, precision and recall of the models were then recorded.

# RESULT

After evaluation of the models on the held-out validation set, the Micronet, ResNet model had an overall accuracy of 61% which is substantially lower than that obtained from the ImageNet Resnet model which had an accuracy of 81%. Similarly, the weighted precision and weighted recall of the Imagenet ResNet model 86% and 81% respectively were also higher than that obtained for the MicroNet ResNet model with 62% and 60% respectively shown in Table 2.

**Table 2: Table of accuracy, precision and recall of the models**

| S/N | Model | Accuracy | Weighted Precision | Weighted Recall | Weighted f1 score |
|---|---|---|---|---|---|
| 1. | Original ResNet | 0.81 | 0.86 | 0.81 | 0.77 |
| 2. | Pretrained MicroNet | 0.60 | 0.62 | 0.60 | 0.58 |

On the individual performance on the different classes, pretrained ImageNet ResNet models significantly performed a lot better with precision values ranging from 0.70 (for tigernut starch) to 1.00 (Green peas and oat ) except for a Cassava starch where the model has a precision of 0.41. The model also has significantly high recall values (> 0.85) for most of the starch samples except for Maize and Millet starch samples which were significantly low (0.23 and 0.31 respectively). However, it was observed that for the 3 classes in which the model didn't perform as great as others for either the precision or recall, a trade-off was made between the precision and recall i.e one of the two was low at a time, not both at the same time.

**Table 3: Table of result of Imagenet Resnet Model showing the Precision, Recall and f1-score**

| Starch sample | Precision | recall | f1_score | support | labels |
|---|---|---|---|---|---|
| Cassava | 0.411765 | 0.933333 | 0.571429 | 15.0 | Imagenet |
| Green peas | 1.000000 | 1.000000 | 1.000000 | 26.0 | Imagenet |
| Irish potato | 0.916667 | 1.000000 | 0.956522 | 22.0 | Imagenet |
| Maize | 0.857143 | 0.230769 | 0.363636 | 26.0 | Imagenet |
| Millet | 0.833333 | 0.312500 | 0.454545 | 16.0 | Imagenet |
| Oat | 1.000000 | 0.937500 | 0.967742 | 16.0 | Imagenet |
| Rice | 0.928571 | 0.866667 | 0.896552 | 15.0 | Imagenet |
| Tigernut | 0.700000 | 0.954545 | 0.807692 | 22.0 | Imagenet |
| Wheat | 0.904762 | 1.000000 | 0.950000 | 19.0 | Imagenet |

For the pretrained MicroNet ResNet models, very good precision values were obtained for Cassava starch, Irish potato starch, Rice starch and wheat starch (1.00, 1.00, 1.00 and 0.79 respectively) others gave a value ranging from 0.56 (Maize starch) to 0.12, Millet starch. The model also has high recall values (>0.7) for 5 out of the 9 different starch samples namely Irish potato starch (0.95), Maize starch (0.84), Rice starch (0.73), Tigernut starch (0.91) and Wheat starch (0.79). However, this model failed at identifying any of the millet starch samples having a precision value of 0.12 and recall value of 0.06.

Table 4: Table of the result of pretrained Micronet Resnet Model showing the Precision, Recall and f1-score

| Starch sample | Precision | recall | f1_score | support | labels |
|---|---|---|---|---|---|
| Cassava | 1.000000 | 0.266667 | 0.421053 | 15.0 | micronet |
| Green peas | 0.500000 | 0.461538 | 0.480000 | 26.0 | micronet |
| Irish potato | 1.000000 | 0.954545 | 0.976744 | 22.0 | micronet |
| Maize | 0.564103 | 0.846154 | 0.676923 | 26.0 | micronet |
| Millet | 0.125000 | 0.062500 | 0.083333 | 16.0 | micronet |
| Oat | 0.285714 | 0.125000 | 0.173913 | 16.0 | micronet |
| Rice | 1.000000 | 0.733333 | 0.846154 | 15.0 | micronet |
| Tigernut | 0.454545 | 0.909091 | 0.606061 | 22.0 | micronet |
| Wheat | 0.789474 | 0.789474 | 0.789474 | 19.0 | micronet |

A more in depth analysis of the pretrained MicroNet ResNet model showed the model misclassified most of the millet starch samples for tigernut starch as shown in the confusion matrix in table 4

**Table 5: Confusion matrix of pretrained MicroNet model with actual class as the column and predicted classes as the row**

| Predicted/Actual | Cassava | Green peas | Irish potato | Maize | Millet | Oat | Rice | Tigernut | Wheat |
|---|---|---|---|---|---|---|---|---|---|
| Cassava | 4 | 6 | 0 | 3 | 0 | 0 | 0 | 2 | 0 |
| Green peas | 0 | 12 | 0 | 0 | 6 | 0 | 0 | 8 | 0 |
| Irish potato | 0 | 0 | 21 | 1 | 0 | 0 | 0 | 0 | 0 |
| Maize | 0 | 3 | 0 | 22 | 0 | 1 | 0 | 0 | 0 |
| Millet | 0 | 1 | 0 | 0 | 1 | 0 | 0 | 14 | 0 |
| Oat | 0 | 1 | 0 | 13 | 0 | 2 | 0 | 0 | 0 |
| Rice | 0 | 0 | 0 | 0 | 0 | 0 | 11 | 0 | 4 |
| Tigernut | 0 | 1 | 0 | 0 | 1 | 0 | 0 | 20 | 0 |
| Wheat | 0 | 0 | 0 | 0 | 0 | 4 | 0 | 0 | 15 |

A comparison of pretrained Imagenet model and Micronet showed that the pretrained Imagenet model significantly outperformed the pertained Micronet on all but 3 (Cassava starch, Irish potato starch and Rice starch) for the precision value as shown in Figure 2. However, while the performance of the Imagenet model is still considerably good (>0.9) of the later two, it significantly performs better for most of the classes.

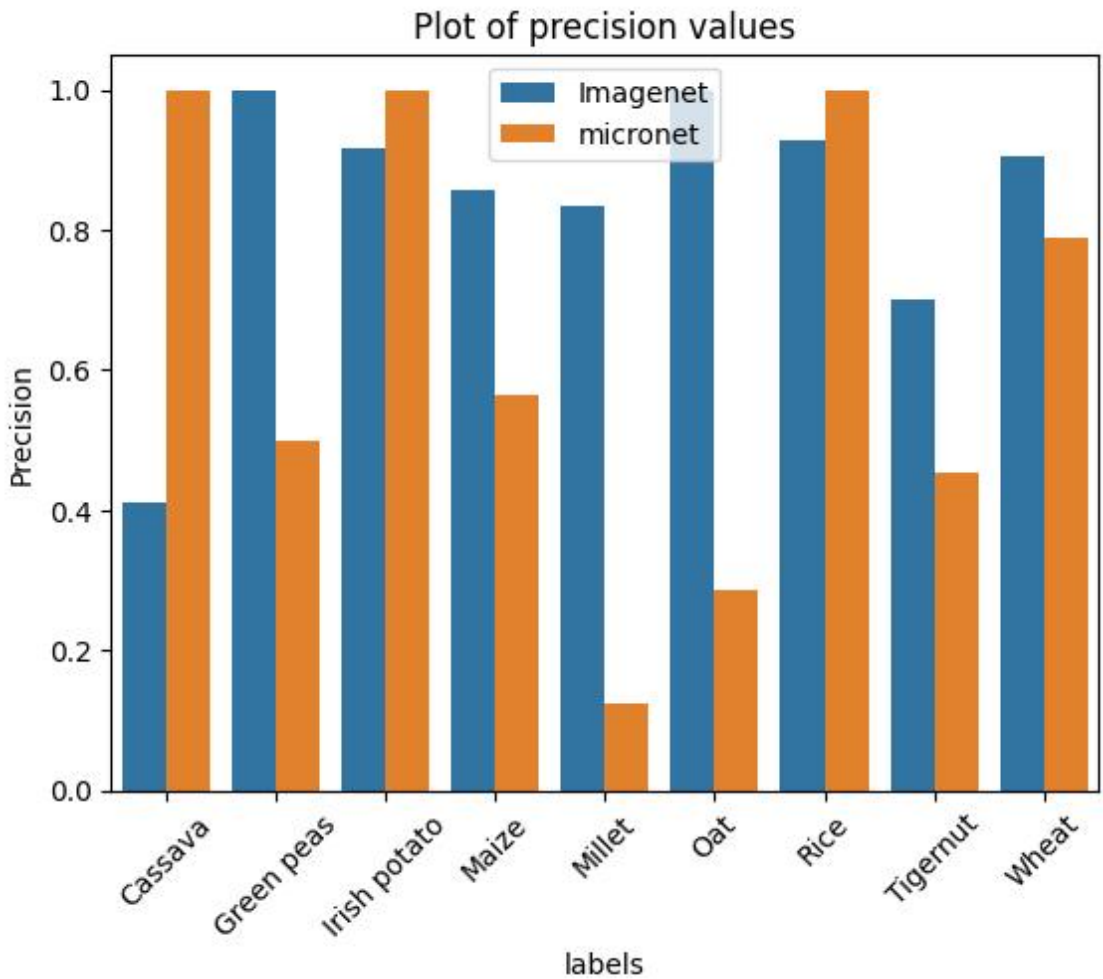

Fig. 3: Plot of the precision value for the Micronet and ResNet model on the different starch samples.

Similarly, a comparison of the recall value for the pretrained Imagenet model and Micronet showed that the pretrained Imagenet model has significantly better performance (> 0.8) on most of the starch images (7 out of 9 classes) compared to the pretrained Micronet model that has a recall value of > 0.8 only in 4 out of 9 classes as shown in Figure 3.

However, the performance of both the pretrained Imagenet model and Micronet were significantly low (<0.4) on the millet starch classification and identification.

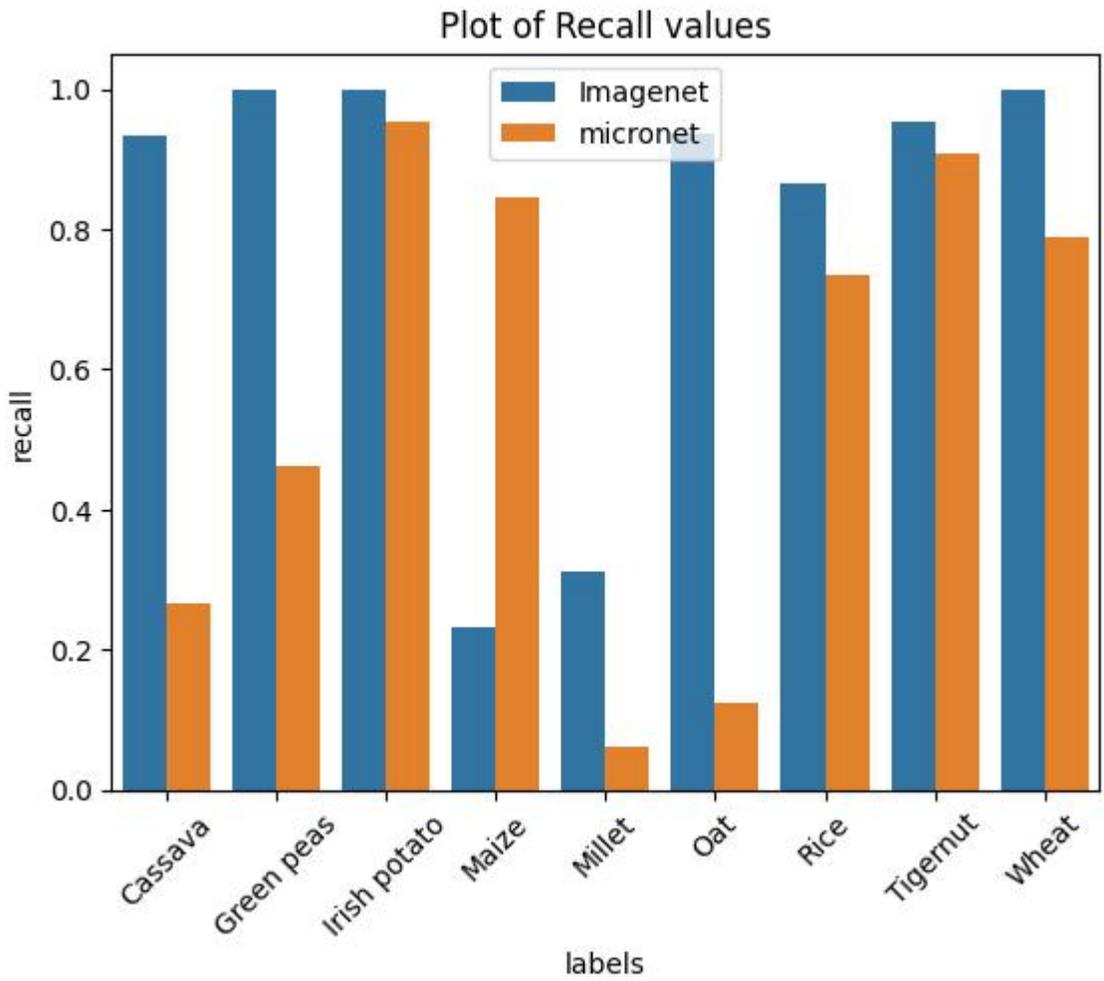

Fig. 4: Plot of the recall value for the Micronet and ResNet model on the different starch sample

**DISCUSSION**

The result of the analysis showed that the fine tuned pretrained ResNet model on the Imagenet dataset has a better performance than the fine tuned pretrained model on the Micronet dataset despite the greater similarity of the large-scale Micronet dataset to the dataset used for the research which both contain microscopic images as compared to the Imagenet dataset that contains random day to day things like dogs, cats, tables etc.

The concept of transfer learning (i.e. training a model initially on a large dataset and then fine tuning on a smaller dataset) was introduced to solve the insufficient dataset to train deep neural networks (Tan et al., 2018).

It was posited that by training initially on the larger dataset, the model will be able to learn common features such as edges, corners etc that it will need to learn also on the smaller dataset, thus, the model can learn the low-level features during the initial pretraining session and more granular features needed for generalization and identification of the specific data during the finetuning and training session.

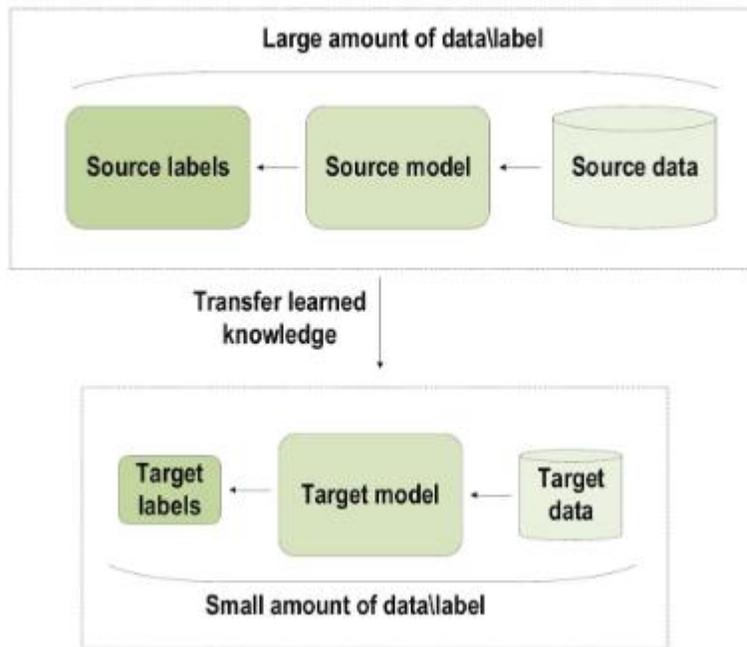

Fig. 5: A conceptual diagram of transfer learning.

Convolution Neural Networks (CNNs), the type of deep learning used in this research that has been previously pretrained on the Imagenet dataset have consistently shown the ability to demonstrate superior classification and segmentation tasks on microscopy (, Akers et al., 2021, Durmaz et al., 2021, Goetz et al., 2022), however given the dissimilarity between most images in the imagenet and most microscopic images, the MicroNet dataset was introduced, containing 100,000 microscopic images for pre-training models specifically for segmentation of the microscopical images as observed by Stuckner et al., (2022).

While the pretrained MicroNet models were designed for segmentation tasks of microscopic image which are tedious, time consuming and prone to bias due to the requirement of manual measurements (ASTM E112, 2010), models for this task usually require a CNN based encoder that is incorporated into the encoder-decoder type architecture for this task. These architectures then benefit from transfer learning involving pretrained encoder models that contains useful learnt information for the other image analysis tasks (Stuckner et al., 2022).

This encoder contain different types of CNN architectures such as ResNet (Szegedy et al., 2017), VGG (Kim et al., 2016), EfficientNet (M. Tan & Le, 2019), DenseNet (Huang et al., 2017) etc. Thus, the ResNet encoder pretrained on the MicroNet and Imagenet dataset were selected.

From a practical understanding, it was expected that the pretrained MicroNet was expected to perform better than that on the Imagenet given that project was a classification task on microscopical images, similar to what the Micronet dataset that the model was pretrained on. However, as posited by Pan & Yang, (2010), transfer learning only works while the learned representation on initial data risk task are applicable to the target and thus fail if the transferred knowledge has negative impact on the target task. While it might be hard to say if the initial task has a negative on the desired task, the divergence between the data source for micronet and the target task could be a potential source of failure given that images for MicroNet were obtained from different datasources mostly that used electron Microscopes (Aversa et al., 2018, DeCost & Holm, 2016, DeCost et al., 2017b), while the dataset for the research were taken with phone cameras from light microscopes showed the task dataset might have more in common with Imagenet dataset than MicroNet dataset.

The precision given by equation 3 shows the number of actual correct images that the models predicted corrected compared to the number of all images the model predicted to be a particular class. This showed for each class how well the model was able to capture the peculiarities of each of the classes, thus showing us the fraction of the actual trues values for all the values the models predicted to each class

$$\text{Precision} = \frac{True\ Positive}{True\ Positive + False\ Positive}$$

$$= \frac{True\ Positive}{Total\ Predicted\ Positive}$$

*Equation 3: equation for precision*

Recall (also known as sensitivity) however tells us the fraction of the correct labels/classes is captured by the predicted classes for the images, as shown in equation

$$recall = \frac{True\ Positives}{True\ Positives + False\ Negatives}$$

*Equation 4: equation for recall*

However, given the tradeoff of model recall and precision (Davis & Goadrich, 2006), the F1 score was introduced to give a singular value that contain information about the fraction of the predicted value/class is actually correct (precision) and fraction of the actual value/class is correctly predicted and is shown in equation 5. This value showed that the Imagenet pretrained model perform significantly better than the MicroNet model.

$$F_1 = 2 * \frac{precision * recall}{precision + recall}$$

*Equation 5: equation of f1 score.*

However, some other important factors that affected the performance of the model include the quantity and quality of data. Given that deep learning models require a lot of data to work efficiently even for transfer learning, getting more data could potentially improve the performance of the models on the dataset. The quality of the data is also another important consideration as the images used for the analysis were obtained by taking snapshots using smartphones from the lens of the microscope, with more qualitative data, it is also expected that the model improves also.

**CONCLUSION**

The application of artificial intelligence methods offers an opportunity to accurately identify and classify starches quickly. The use of pretrained model on larger datasets can also be used to augment small data sizes to get very good results. In addition for this task, while the dataset are microscopical images of starch samples obtained from the lab, model pretrained on Imagenet seem to perform a lot better than that pretrained on microscopical images from MicroNet. Future directions for this work will be to benchmark the accuracy, precision and recall of these AI models against human experts, which is the current global used method for identification of starch samples from their microscopic images given the domain knowledge required to be able to perform this task. Another future direction for this project is the segmentation, identification and quantification of each individual starch sample in the microscopical image for purity estimation of the starch samples. This is because beyond the homogenous exhibition of the microscopical images, there exist the possibility of mixup of multiple and different starch samples inside the same microscopical image, thus the ability to segment and identify the different starch samples in the microscopical image and also estimate the quantity of this different starch samples will be of great use for the practical implementation of this idea.